\documentclass[conference]{IEEEtran}
\IEEEoverridecommandlockouts

\usepackage{cite}
\usepackage{amsmath,amssymb,amsfonts}
\usepackage{algorithmic}
\usepackage{fancyhdr}
\usepackage{graphicx}
\usepackage{lipsum}
\usepackage{textcomp}
\usepackage{xcolor}
\usepackage{multirow}
\usepackage{enumerate}
\usepackage{booktabs}
\usepackage{caption}
\usepackage{subcaption}
\usepackage{float}
\usepackage{comment}

\def\bng{\bngx}

\font\bngx=bang10

\def\*#1*#2{o\null{#2}{#1}}

\def\sh#1{\setbox0=\hbox{#1}%
     \kern-.02em\copy0\kern-\wd0
     \kern.04em\copy0\kern-\wd0
     \kern-.02em\raise.0433em\box0 }
\usepackage[colorlinks, linkcolor=blue, citecolor=blue, urlcolor=blue]{hyperref}

\usepackage{array}
\usepackage{balance}
\def\BibTeX{{\rm B\kern-.05em{\sc i\kern-.025em b}\kern-.08em
    T\kern-.1667em\lower.7ex\hbox{E}\kern-.125emX}}

\fancypagestyle{firstpage}
{
    \fancyhead[R]{} 
    \fancyhead[L]{2024 27th International Conference on Computer and Information Technology (ICCIT),\\ 20-22 December 2024, Cox’s Bazar, Bangladesh}
    \fancyfoot[C]{} 
    \fancyfoot[L]{979-8-3315-1909-4/24/\$31.00 ©2024 IEEE}
    \setlength{\footskip}{25pt} 
}

\begin{document}

\makeatletter
    \newcommand{\linebreakand}{%
      \end{@IEEEauthorhalign}
      \hfill\mbox{}\par
      \mbox{}\hfill\begin{@IEEEauthorhalign}
    }
\makeatother

\title{Bridging Dialects: Translating Standard Bangla to Regional Variants Using Neural Models}

\author{\IEEEauthorblockN{ Md. Arafat Alam Khandaker\IEEEauthorrefmark{1}, Ziyan Shirin Raha\IEEEauthorrefmark{1}, Bidyarthi Paul\IEEEauthorrefmark{1},   Tashreef Muhammad\IEEEauthorrefmark{1}
}
\IEEEauthorblockA{\IEEEauthorrefmark{1}Department of Computer Science and Engineering, Ahsanullah University of Science and Technology,\\ Dhaka, Bangladesh}

aarafatalam18@gmail.com, ziyanraha@gmail.com, 
bidyarthipaul01@gmail.com, tashreef.muhammad@seu.edu.bd 
}

\maketitle
\thispagestyle{firstpage}

\begin{abstract}
The Bangla language includes many regional dialects, adding to its cultural richness. The translation of Bangla Language into regional dialects presents a challenge due to significant variations in vocabulary, pronunciation, and sentence structure across regions like Chittagong, Sylhet, Barishal, Noakhali, and Mymensingh. These dialects, though vital to local identities, lack of representation in technological applications. This study addresses this gap by translating standard Bangla into these dialects using neural machine translation (NMT) models, including BanglaT5, mT5, and mBART50. The work is motivated by the need to preserve linguistic diversity and improve communication among dialect speakers. The models were fine-tuned using the ``Vashantor'' dataset, containing 32,500 sentences across various dialects, and evaluated through Character Error Rate (CER) and Word Error Rate (WER) metrics. BanglaT5 demonstrated superior performance with a CER of 12.3\% and WER of 15.7\%, highlighting its effectiveness in capturing dialectal nuances. The outcomes of this research contribute to the development of inclusive language technologies that support regional dialects and promote linguistic diversity.
\end{abstract}


\section{Introduction}
\label{sec:introduction}
Bangladesh is a country rich in linguistic diversity, with over 50 languages spoken across its 64 districts. Within the Bangla language itself, there are numerous regional dialects that vary significantly from one another in terms of vocabulary, pronunciation, tone, and sentence structure. These dialects, spoken in regions like Chittagong, Sylhet, Barishal, and Mymensingh, reflect the cultural and social uniqueness of each area. However, despite their prevalence, these dialects are often underrepresented in both academic research and technological applications. This study aims to bridge this gap by focusing on the translation of standard Bangla into these regional dialects using advanced neural machine translation (NMT) models.

Neural Machine Translation (NMT)~\cite{NMT} is a method for translating words from one language to another. It is an innovative approach to machine translation that employs a single, massive neural network trained end-to-end and organized around an encoder-decoder architecture. The encoder converts a source sentence to a continuous space representation, and the decoder generates the translation from this representation.

The impulsion for this research arises from the clear lack of studies dedicated to translating standard Bangla into its regional variants. Previous research has primarily concentrated on translating Bangla into other languages, such as English and Hindi, or on mixed languages like Banglish. Although in the study~\cite{2} they translated regional to Bangla language, there is a noticeable void in the reverse process—translating standard Bangla into its regional dialects. This gap is significant because it limits the understanding and accessibility of these dialects, which are an integral part of the cultural identity of their speakers. By addressing this gap, the study seeks to contribute to the preservation and promotion of linguistic diversity within Bangladesh.

To tackle the challenge of translating standard Bangla into regional dialects, the study employs several cutting-edge neural models, including BanglaT5\cite {t5}, mT5\cite {mt5}, and mBART50 \cite {mbart50}. These models are fine-tuned specifically to handle the linguistic complexities of the selected dialects. The approach involves training these models on a dataset that includes extensive samples of standard Bangla and its regional variants.
The models leverage continuous representations and attention mechanisms, enabling them to focus on relevant parts of the input sentence and manage long-distance dependencies. By fine-tuning these models, the study aims to produce translations that are not only linguistically accurate but also culturally and contextually appropriate. The performance of these models is evaluated using established metrics like Character Error Rate (CER) \cite{cer} and Word Error Rate (WER) \cite{wer}. WER evaluates word-level accuracy by measuring the number of insertions, deletions, and substitutions in the translation, while CER captures more granular, character-level errors, critical for preserving linguistic details. The contribution of the research paper can be summarized as follows:

\begin{itemize}
  \item  The research fills a significant gap in the field of translation studies by addressing the underexplored area of translating standard Bangla into regional dialects, a topic that has received limited attention in prior work. While earlier research has focused on translating Bangla into other languages or translating regional dialects into standard Bangla, this study uniquely contributes to the reverse process.

  \item We fine-tuned several neural models, including BanglaT5, mT5, and mBART50, to translate standard Bengali into various regional dialects. We afterwards evaluated the performance of these models across different dialects to identify their strengths and areas for improvement.
\end{itemize}
This study begins with an introduction in Section \ref{sec:introduction}, followed by a review of related works in Section \ref{sec:Related Works}. The methodology is detailed in Section \ref{sec:methodology}, results are analyzed in Section \ref{sec:result analysis}, and limitations with future work are discussed in Section \ref{sec:limitation}. Finally, Section \ref{sec:conclusion} summarizes the study.

\section{Related Works}
\label{sec:Related Works}
This section provides a compact overview of previous study which is relevant to our study, organized into three subsections: Transformer Based Neural Machine Translation, Adversarial Neural Machine Translation and  Sequence-to-Sequence (Seq2Seq) Neural Machine Translation.

\subsection { Transformer-Based Neural Machine Translation}
 Fatema et al. \cite{2} investigates the translation of Bangla regional dialects into standard Bangla using models like Bangla-bert-base, mT5 and BanglaT5. While BanglaT5  performs better in character error rate, word error rate, BLEU, and METEOR scores, Bangla-bert-base surpasses mBERT and mT5 in precision, recall, and F1-score . This work highlighted the superiority of specialized models in dialect translation over general-purpose models. Soran et al. \cite{8} employs transformer-based NMT model for low-resource languages like Kurdish Sorani. It utilizes data from different  sources and attention mechanisms . The model significantly enhances its translation performance and  achieving a BLEU score of 0.45 for accurate translations. Laith et al.\cite{9} proposes a novel Transformer-Based NMT model tailored for Arabic dialects, to solve  low-resource language challenges like unique word order and limited vocabulary, resulting in higher BLEU scores. Significant advancement was demonstrated by extensive testing using several Arabic dialects and translations to Modern Standard Arabic (MSA). 

\subsection{ Adversarial Neural Machine Translation}
 Wenting et al.\cite{10} tackles short sequence translation from Chinese to English using a generative adversarial network (GAN). The GAN has a generator that mimics human translations and a discriminator that tells them apart from real human translations. During training, dynamic discriminators and static BLEU score targets guide the generator. Tests on an English-Chinese dataset showed this method improved translation quality by over 8\% compared to typical recurrent neural network (RNN) models, achieving an average BLEU score of 28.2. Lijun et al. \cite{11} presentes Adversarial-NMT, a new neural machine translation (NMT) approach. Instead of just mimicking human translations, Adversarial-NMT uses a CNN adversary to minimize differences between human and NMT translations. The NMT model and adversary are co-trained using a policy gradient method. Tests on English-French and German-English tasks showed that Adversarial-NMT significantly improved translation quality over strong baseline models. Adversarial-NMT translated about 59.4\% of sentences better than the baselines.

\subsection{Sequence-to-Sequence (Seq2Seq) Neural Machine Translation }
 The exploration of machine translation between English and Bangla using Recurrent Neural Networks (RNNs) is discussed in first two papers. Rafiqul et al.\cite{12} uses a Seq2Seq model with attention-based RNNs and cross-entropy loss metrics, achieving less than 2\% loss with a dataset of over 6,000 Bangla-English sentence pairs. Shaykh et al.\cite{13} employs an encoder-decoder RNN with a knowledge-based context vector, utilizing 4,000 parallel sentences. They addressed the problem of different sentence lengths by using linear activation in the encoder and tanh activation in the decoder, which led to the best outcomes, found that Gated Recurrent Units (GRUs) outperform Long Short-Term Memory (LSTM) networks and emphasized the importance of attention mechanisms with softmax and sigmoid activations. Arid et al.\cite{14}  investigates Neural Machine Translation (NMT) methods for translating between Bangla and English. It evaluated Bidirectional Long Short Term Memory (BiLSTM) and Transformer models using datasets like Indic Languages Multilingual Parallel Corpus (ILMPC) and Six Indian Parallel Corpora (SIPC).  The BiLSTM model captured long-term dependencies well, while the Transformer model handled long-term dependencies and parallelization efficiently.\\
Table~\ref{table:tab1}, represents a summary of existing works that were conducted. It provides a comprehensive outlook on the related works in the domain.
 
\begin{table*}[h]
\caption{A Summary of Existing Works in Machine Translation}
\label{table:tab1}
\normalsize
\centering
\begin{tabular}{|>{\centering\arraybackslash}p{2.5cm}|>{\raggedright\arraybackslash}p{3cm}|>{\centering\arraybackslash}p{0.8cm}|>{\raggedright\arraybackslash}p{10cm}|}
\hline
\textbf{Authors} & \textbf{Use Models} & \textbf{Year} & \textbf{Contribution} \\ \hline

Fatema et al.\cite{2} & mT5, BanglaT5, mBERT, Bangla-bert-base & 2023 & Evaluated the mT5 and BanglaT5 models for translating regional Bangla dialects into standard Bangla and analyzed mBERT and Bangla-bert-base models for identifying regional origins \\ \hline

Soran et al.\cite{8} & Transformer-based Neural Machine Translation Model & 2023 & Developed a transformer-based neural machine translation model for the low-resource Kurdish Sorani dialect, demonstrating high translation quality \\ \hline

Rafiqul et al.\cite{12} & Sequence-to-Sequence Learning Model, Recurrent Neural Networks & 2023 & Focused on Bengali to English translation, addressing the complexities of Bengali syntax and vocabulary \\ \hline

Wenting et al.\cite{10} & Generative Adversarial Network (GAN) & 2022 & Implemented a Generative Adversarial Network (GAN) for Chinese to English translation, which exceeded the performance of Recurrent Neural Networks by 8\% in translation quality on an English-Chinese dataset \\ \hline

Laith et al.\cite{9} & Transformer-Based NMT, Subword Units & 2021 & Introduced a Transformer-Based NMT model specifically designed for Arabic dialects, addressing challenges in low-resource languages through the utilization of subword units \\ \hline

Shaykh et al.\cite{13} & Encoder-Decoder RNN, Knowledge-based Context Vector & 2021 & Focused on translating from English to Bangla using an encoder-decoder RNN architecture with a knowledge-based context vector to enhance translation accuracy \\ \hline

Arid et al.\cite{14} & Neural Machine Translation Techniques & 2019 & Explored various techniques in neural machine translation for the Bangla-English language pair, achieving improvements in translation \\ \hline

Lijun et al.\cite{11} & CNN Adversary, Policy Gradient Co-Training & 2018 & Introduced a CNN adversary and a policy gradient co-training technique \\ \hline

\end{tabular}
\label{table:translation}
\end{table*}

\section{Methodology}
\label{sec:methodology}
This section of the paper describes the dataset that was used on it and the proposed methodology  of translating standard Bengali language to  correlate with different regional dialects. We have split it down in some steps as it displays in Fig \ref{fig:Proposed methodology}. 

\subsection{Dataset} The ``Vashantor" dataset \cite{2} is a significant resource in computational linguistics, especially for the Bangla language. It aids in translating regional Bangla dialects into standard Bangla. The dataset includes 32,500 lines in Bangla, Banglish (a mix of Bangla and English), and English, focusing on five major regional dialects: Chittagong, Noakhali, Sylhet, Barishal, and Mymensingh. These dialects were selected due to their notable differences, which often lead to communication challenges, highlighting the need for effective translation models.

Each entry in the dataset includes a text instance in the regional dialect, a translation into standard Bangla, a transcription in Banglish (translated Bangla), and an English translation for non-native scholars. For example, a sample text from the Chattogram region in the dataset reads: ``{\bng tuim ik kha{I}ch?}", translated to standard Bangla as ``{\bng tuim ik ekhJech?}" and into English as ``What have you eaten?" This uniformed approach is simple assembly to integrate with machine learning frameworks, allowing researchers to train and test translating models effectively.

\subsection{Proposed Methodology} 

\setcounter{figure}{1}
\begin{figure*}[t]
\centering
\includegraphics[width=1.0\textwidth]{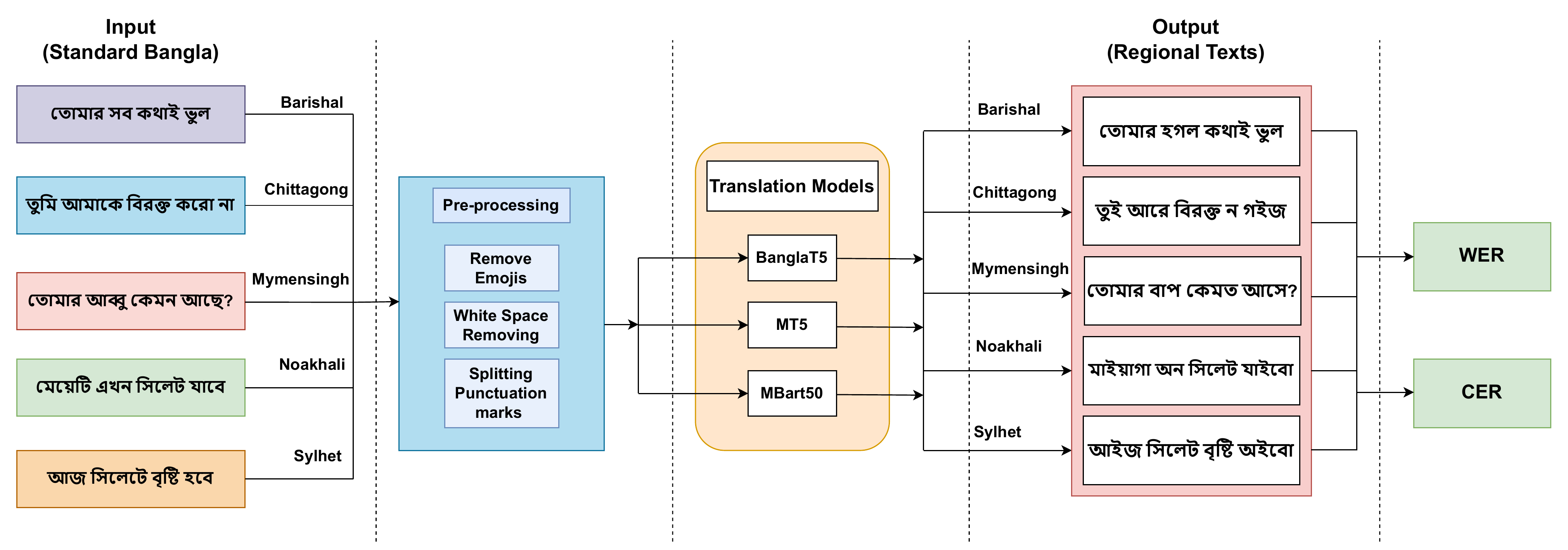}
\caption{Comprehensive Framework for Methodology}
\label{fig:Proposed methodology}
\end{figure*}

The procedures specified for transforming standard Bengali into regional languages. This technique delivers three alternative translation options for each input text. Furthermore, the quality of these translation results is assessed by comparing them to the reference translation. The essential stages of the proposed translation procedure include processing five sentences in various regional dialects.

\textbf{Step i) The input text:} The input text is standard Bangla text. Each text sample is collected and processed sequentially to ensure that the model thoroughly assesses each individual entry. 

\textbf{Step ii) Data pre-processing:} Essential preprocessing tasks include removing irrelevant emojis that introduce noise, meticulously managing whitespace (spaces, tabs, line breaks) to ensure consistent structure, and stripping punctuation. Some related examples are: ``{\bng tuim ik kera?}" to ``{\bng tuim ik kera}", ``{\bng ikH! ekn igeyech {O}khaen?}" to ``{\bng ikH ekn igeyech {O}khaen}". ``{\bng itin Aamar saeth edkha kreln.}" to ``{\bng itin Aamar saeth edkha kreln}"

\textbf{Step iii) Dialect Translation Model:} During this step, we train translation models that are unique to each regional dialect's linguistic features. To translate from Bangla to  dialect, we use models like ``mt5-small" ,``BanglaT5 ", ``mBart-large-50 ". These models are specifically designed to understand the variety and complexity of five regional dialects, resulting in accurate and contextually appropriate translations. 

\subsubsection{mt5-small}
mT5 Small is an optimized version of the Multilingual T5 model that comes pre-trained in over 100 languages, including Bangla, making it a reliable choice for multilingual translation tasks. 

Its multilingual training enables knowledge transfer from high-resource to low-resource languages, hence improving its performance in languages such as Bangla. The efficient architecture and vast pretraining corpus ensure that it can handle a wide range of linguistic structures, resulting in improved translation quality. A study \cite{two} emphasizes multilingual performance across over 100 languages, it also shows the efficiency of mT5 in low-resource languages such as Bangla. In this instance, mT5 produces competitive translation outcomes, which is especially important when computational economy is essential.

\subsubsection{BanglaT5}
Bangla T5 is a fine-tuned version of the T5 framework developed for Bangla-language tasks. By determining translation as a text-to-text concern, Bangla T5 takes advantage of bidirectional encoding for handling the complexity of both source and target sentences making it highly contextual. Its pretraining on Bangla data demonstrates optimal performance in low-resource environments by capturing the syntactic and semantic complexities of Bangla, which are critical for providing grammatically correct translations. This makes it a very useful model for Bangla-specific translation problems. In \cite{one} BanglaT5 model surpassed multilingual models in Bangla-English translation, scoring 38.8 on the SacreBLEU scale. It showed a considerable improvement in machine translation tasks.

\subsubsection{mBart-large-50}
mBART Large 50 is a reducing autoencoder developed for sequence-to-sequence tasks and trained in 50 languages, including Bangla. Its pretraining aim of recreating original texts from altered inputs enhances its ability to produce fluent and coherent translations. The tremendous model size allows for the modeling of complex linguistic structures. The multilingual pretraining permits cross-lingual transfer, making mBART especially useful for low-resource language pairs such as Bangla. 
The mBART-50 model enhances multilingual machine translation effectiveness significantly, especially for low-resource languages, by up to 12 BLEU points in these conditions and more than 5 BLEU points for document-level and unsupervised tasks. \cite{three} Its eliminating autoencoder pretraining on a large multilingual corpus provides strong performance across multiple language pairs, including those omitted from the pretraining data.

\textbf{Step iv) Hyperparameter Tuning:} We aim to identify the optimal combination to enhance model performance. Therefore, we experiment with different combinations by fine-tuning the hyperparameters. Various learning rates ranging from 1e-3 to 1e-5 were used to train the model, with each experiment utilizing batch sizes between 8 and 16. Additionally, the experiments were conducted over 30 to 50 epochs with the dataset. This comprehensive approach allowed us to explore the interplay between batch size, learning rate, and the number of epochs, ensuring a thorough evaluation of the model's performance.

\textbf{Step v) Generation of Translation:} For each input sample, we generate three alternative translations using three different models Bangla T5, MT5, and mBART-50. This gives us fifteen different translations for the five regional dialects. This method allows us to assess numerous translation possibilities and select the one that best represents the standard Bangla language. 

\textbf{Step vi) Translation Quality Assessment:} We use two types of assessment metrics to assess the quality of Bangla to dialect translation. These indicators help us understand the translation's accuracy and consistency. We employed CER and WER metrics to assess translation quality. In the result analysis section, we will discuss the score they received.

\section{Result Analysis}
\label{sec:result analysis}

\begin{table*}[htbp]
\centering
\caption{CER, WER scores of all the Bangla regional dialect translation models}
\scalebox{1.15}{
\begin{tabular}{|c|c|cc|cc|cc|cc|cc|}
\hline
         \textbf{Evaluation}            & \multirow{2}{*}{\textbf{Models}} & \multicolumn{2}{c|}{\textbf{Barishal}}                             & \multicolumn{2}{c|}{\textbf{Noakhali}}                 & \multicolumn{2}{c|}{\textbf{Mymensingh}}               & \multicolumn{2}{c|}{\textbf{Sylhet}}                   & \multicolumn{2}{c|}{\textbf{Chittagong}}               \\ \cline{1-1} \cline{3-12} 
Epoch                &                         & \multicolumn{1}{c|}{30}     & 50                          & \multicolumn{1}{c|}{30}     & 50              & \multicolumn{1}{c|}{30}     & 50              & \multicolumn{1}{c|}{30}     & 50              & \multicolumn{1}{c|}{30}     & 50              \\ \hline
\multirow{3}{*}{WER} & mT5                     & \multicolumn{1}{c|}{0.4569} & \multicolumn{1}{l|}{0.3958} & \multicolumn{1}{c|}{0.6974} & 0.5804          & \multicolumn{1}{c|}{0.4265} & 0.3351          & \multicolumn{1}{c|}{0.7956} & 0.6602          & \multicolumn{1}{c|}{0.9264} & 0.7557          \\ \cline{2-12} 
                     & BanglaT5                & \multicolumn{1}{c|}{0.5755} & 0.3932                      & \multicolumn{1}{c|}{0.7065} & \textbf{0.5775} & \multicolumn{1}{c|}{0.3678} & \textbf{0.2380} & \multicolumn{1}{c|}{0.6956} & \textbf{0.6064} & \multicolumn{1}{c|}{0.7408} & \textbf{0.7066} \\ \cline{2-12} 
                     & mBART50                 & \multicolumn{1}{c|}{0.4263} & \textbf{0.3350}             & \multicolumn{1}{c|}{0.7569} & 0.6665          & \multicolumn{1}{c|}{0.3987} & 0.2983          & \multicolumn{1}{c|}{0.8975} & 0.7579          & \multicolumn{1}{c|}{0.9874} & 0.8264          \\ \hline
\multirow{3}{*}{CER} & mT5                     & \multicolumn{1}{c|}{0.3156} & 0.2288                      & \multicolumn{1}{c|}{0.4123} & 0.2949          & \multicolumn{1}{c|}{0.3126} & 0.2017          & \multicolumn{1}{c|}{0.4866} & \textbf{0.3326} & \multicolumn{1}{c|}{0.6812} & 0.4218          \\ \cline{2-12} 
                     & BanglaT5                & \multicolumn{1}{c|}{0.3769} & 0.2378                      & \multicolumn{1}{c|}{0.4236} & \textbf{0.2952} & \multicolumn{1}{c|}{0.2659} & \textbf{0.1216} & \multicolumn{1}{c|}{0.3847} & 0.3345          & \multicolumn{1}{c|}{0.4045} & \textbf{0.3706} \\ \cline{2-12} 
                     & mBART50                 & \multicolumn{1}{c|}{0.2365} & \textbf{0.1789}             & \multicolumn{1}{c|}{0.5894} & 0.4014          & \multicolumn{1}{c|}{0.3145} & 0.1749          & \multicolumn{1}{c|}{0.6164} & 0.4289          & \multicolumn{1}{c|}{0.7523} & 0.4896          \\ \hline
\end{tabular}
\label{tab:scores}
}
\end{table*}

\subsection{Baseline Setup}

The trials were carried out using two separate configurations. The initial setup included Google Colaboratory, Python 3.10.12, PyTorch 2.0.1, a Tesla T4 GPU (15 GB), 13.5 GB of RAM and 68 GB of storage. The second setup also uses Google Colaboratory, Python 3.10.12, PyTorch 2.0.1 and a Tesla A100 GPU.

\begin{figure}
    \centering
    \begin{subfigure}[b]{0.48\textwidth}
        \centering
        \includegraphics[width=\textwidth]{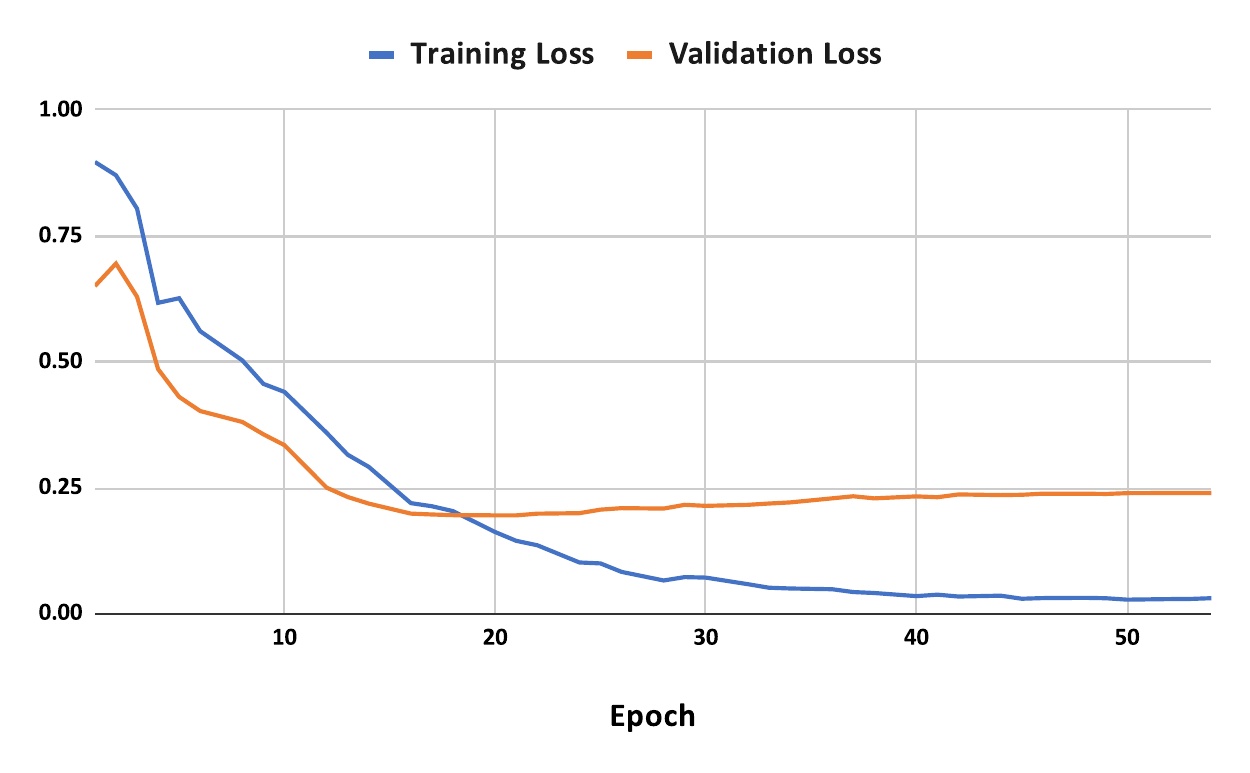}
        \caption{mT5}
        \label{fig3}
    \end{subfigure}
    \hfill
    \begin{subfigure}[b]{0.48\textwidth}
        \centering
        \includegraphics[width=\textwidth]{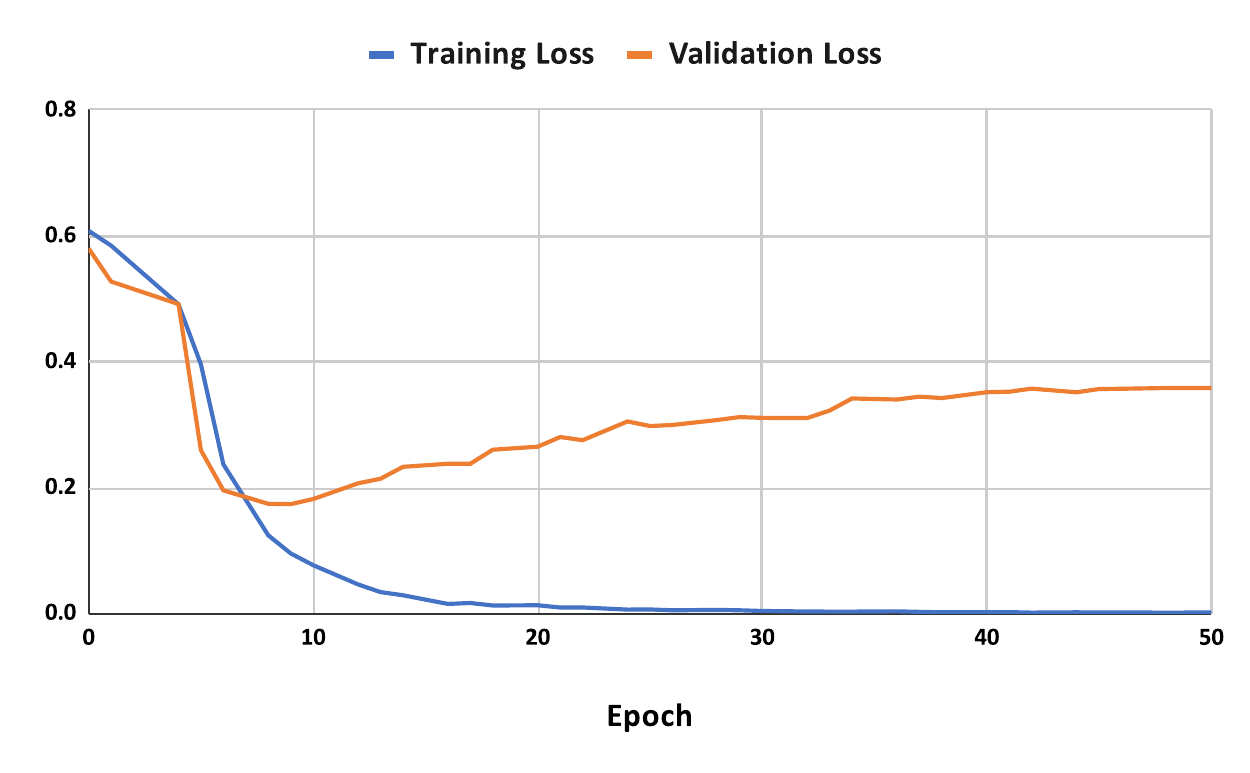}
        \caption{BanglaT5}
        \label{fig4}
    \end{subfigure}
    \vfill
    \begin{subfigure}[b]{0.48\textwidth}
        \centering
        \includegraphics[width=\textwidth]{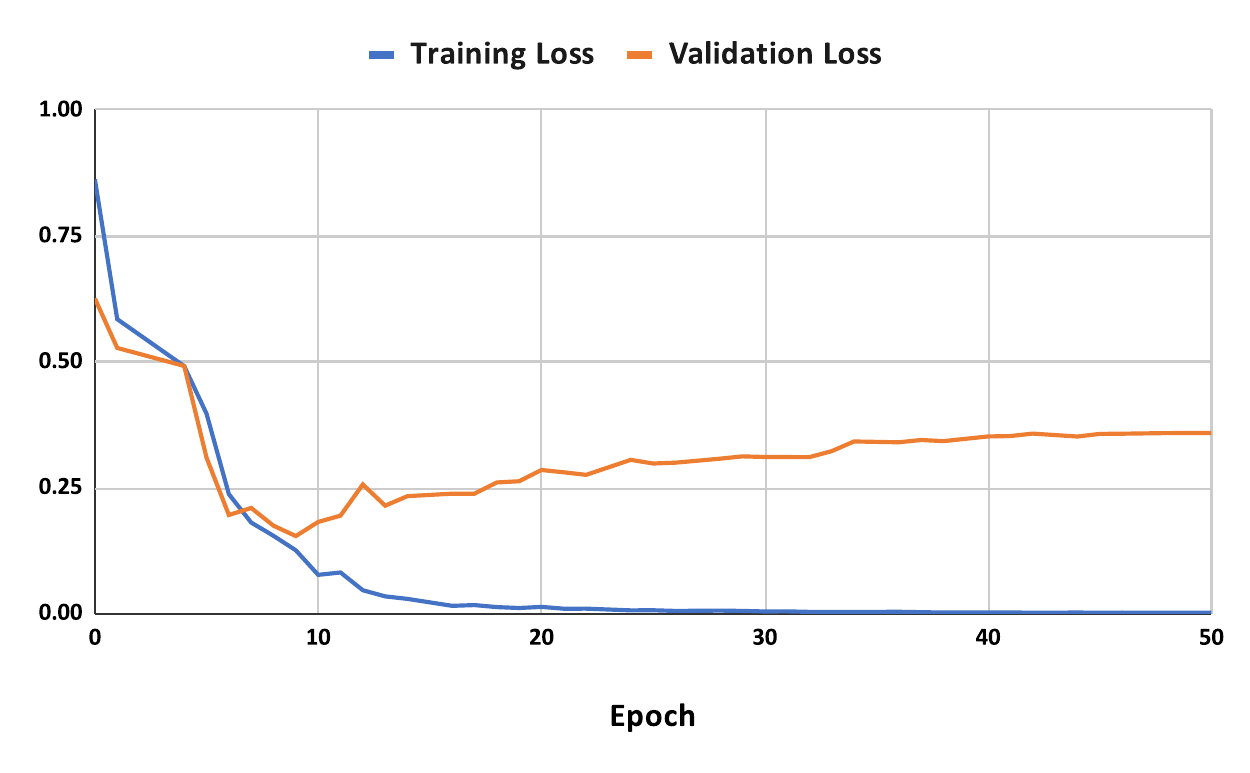}
        \caption{mBART50}
        \label{fig5}
    \end{subfigure}
    \hfill
    \caption{Training Loss vs Validation Loss}
    \label{fig:figloss}
\end{figure}

\subsection{Experimental Result}

This research analyzes the performance of three transformer models—mBART, BanglaT5 and mT5—for translating Bangla regional languages. The models were trained and validated using datasets that included multiple dialect regions, and their performance was measured using metrics such as Word Error Rate (WER), Character Error Rate (CER).

Table ~\ref{tab:scores} contrasts the performance metrics—Character Error Rate (CER) and Word Error Rate (WER)—of three distinct neural machine translation models, mT5, BanglaT5, and mBART50, across several Bangla regional dialects. Each model was evaluated with five different dialects: Barishal, Noakhali, Mymensingh, Sylhet, and Chittagong.

Figure ~\ref{fig:figloss}, and it's sub-figures as Figure ~\ref{fig3}, ~\ref{fig4} and ~\ref{fig5}, represents the training and validation loss curve of mT5, BanglaT5 and mBART50.
The investigation of the BanglaT5, mT5, and mBART50 models toward several Bangla dialects demonstrates significant performance diversity, reflecting the dialect's inherent linguistic complexity. The models were assessed using Word Error Rate (WER) and Character Error Rate (CER) metrics, that provided information on their translation accuracy.BanglaT5 performed effectively in the Mymensingh dialect, with the lowest WER of 23.81\% and CER of 12.16\%, indicating the ability for this dialect due to effective learning. However, it struggled with the Sylhet and Chittagong dialects, recording higher error rates of 64.68\% WER and 33.46\% CER, and 74.65\% WER and 43.06\% CER, respectively, indicating problems in handling these more complicated or underrepresented dialects.
mT5 performed comparably to BanglaT5 in the Barishal and Sylhet dialects, but exhibited a little enhancement in the Noakhali dialect with a WER of 58.05\% and a CER of 29.49\%. Although it struggled with the complicated Sylhet and Chittagong dialects. 
mBART50 superior to both models in the Barishal dialect, with a WER of 33.51\% and a CER of 17.90\%, but lagged in the Noakhali and Chittagong dialects, with significantly higher error rates. This pattern implies that, despite its capabilities, mBART50 requires additional refinements to adequately handle the linguistic aspects of more difficult dialects. These findings show the potential of transformer models in regional language translation, but also the need for further model tuning and dataset optimization to more effectively manage dialectal diversity.

\section{Limitation and Future work}
\label{sec:limitation}
The significant error rates for the Chittagong and Sylhet dialects reflect possible marginalization in the training data, as well as limits in the current transformer models' ability to completely capture the complicated linguistic nuances of these languages. In addition, the intrinsic phonetic and semantic intricacies of the Bangla language provide issues that current preprocessing methods may not fully handle. Future research will concentrate on augmenting training datasets with a broader representation of regional dialects and investigating new transformer topologies designed for multilingual translation issues. Improvements in preprocessing strategies to better manage dialect-specific characteristics, as well as rigorous validation frameworks, will be sought to improve model performance and practicality.

\balance

\section{Conclusion}
\label{sec:conclusion}
The research utilizes advanced neural machine translation models, such as mBART50, BanglaT5 and mT5 to tackle the translation of mainstream Bengali into regional languages. Using the ``Vashantor" dataset, our findings revealed the intricacies of dialectal translation, with large performance differences between dialects. Notably, the Mymensingh dialect had the lowest mistake rates, implying greater representation or translatability in the dataset, whereas the Chittagong and Sylhet dialects presented significant issues, highlighting the need for more specialized techniques and improved training data. Although our trials demonstrate improvement in dialectal translation, they also highlight the importance of training data quality and the possible limitations of current measures in capturing dialect nuances. Future research should look into more diversified model architectures and datasets, as well as novel evaluation methodologies, to improve translation accuracy and robustness across dialects.

\end{document}